\documentclass[onefignum,onetabnum]{siamonline190516}

\usepackage[noadjust]{cite}
\usepackage{xcolor}
\usepackage{graphicx}
\usepackage{algorithm}
\usepackage{amsmath}
\usepackage{amssymb}
\usepackage{hyperref}
\usepackage{textcomp}
\usepackage{url}

\hypersetup{
    colorlinks,
    citecolor=green,
    filecolor=black,
    linkcolor=blue,
    urlcolor=blue
}

\usepackage{xifthen}

\usepackage{algorithm,algcompatible}

\usepackage{xcolor}

\def\clap#1{\hbox to 0pt{\hss#1\hss}}

\def\mathclap{\mathpalette\mathclapinternal}

\def\mathclapinternal#1#2{%
\clap{$\mathsurround=0pt#1{#2}$}}

\usepackage{tikz}

\usetikzlibrary{arrows,fadings,patterns,shapes}
\usetikzlibrary{decorations.pathmorphing}
\usetikzlibrary{decorations.pathreplacing}

\definecolor{PlotColorA}{HTML}{1f77b4}
\definecolor{PlotColorB}{HTML}{ff7f0e}
\definecolor{PlotColorC}{HTML}{2ca02c}
\definecolor{PlotColorD}{HTML}{d62728}
\definecolor{PlotColorE}{HTML}{9467bd}
\definecolor{PlotColorF}{HTML}{8c564b}
\definecolor{PlotColorG}{HTML}{e377c2}
\definecolor{PlotColorH}{HTML}{fce94f}

\newcommand{\dx}{\mathit{dx}}
\newcommand{\dt}{\mathit{dt}}

\renewcommand{\div}{\operatorname{div}}
\newcommand{\grad}{\nabla}

\newcommand{\vv}[1]{\mathbf{#1}}

\begin{document}

\title{Solving Image PDEs with a Shallow Network}

\author{Pascal Getreuer\thanks{Google Research, Mountain View CA, USA} \and Peyman Milanfar\footnotemark[\value{footnote}] \and  Xiyang Luo\footnotemark[\value{footnote}]}

\markboth{
}{Getreuer \MakeLowercase{\textit{et al.}}:
Solving Image PDEs with a Shallow Network}

\maketitle

\begin{abstract}
Partial differential equations (PDEs) are typically used as models of physical
  processes but are also of great interest in PDE-based image processing.
  However, when it comes to their use in imaging, conventional numerical methods
  for solving PDEs tend to require very fine grid resolution for stability, and
  as a result have impractically high computational cost. This work applies
  BLADE  (Best Linear Adaptive Enhancement), a shallow learnable filtering
  framework, to PDE solving, and shows that the resulting approach is efficient
  and accurate, operating more reliably at coarse grid resolutions than
  classical methods. As such, the model can be flexibly used for a wide variety
  of problems in imaging.
\end{abstract}


\section{Introduction}

Classically, finite element numerical schemes for solving PDEs
are derived from local Taylor series analysis, essentially approximating the
solution to be locally a polynomial (see for instance \cite{lele1992compact}).
An important requirement is that the scheme
is stable, meaning that the approximation error accumulated over multiple
successive time steps stays under control instead of exploding. Classical
schemes tend to require that both the time step and grid step are very small in
order to ensure stability. There are numerous techniques to do it better, and
accuracy and stability requirements depend a lot on the particular PDE at hand,
but generally the trade-off between stability and step size is the key difficulty.

BLADE is a trainable adaptive filtering framework that is simple, fast, and
useful for a wide range of imaging problems. In contrast to classic finite
difference solvers, BLADE is a generic data-driven method. A training set is
first formed from a set of input images and corresponding target outputs, which
can be computed once by a more computationally intensive reference method. BLADE is then trained to
approximate the input-target relationship. We previously
showed~\cite{getreuer2018blade} that BLADE is capable of approximating a variety of operators. In this work we show that compelling quality and efficiency is broadly possible on PDEs.

\textbf{Previous work.}
There are two camps of related literature at the intersection of PDE and ML. The first is incorporating ML-models as a part of a PDE solver, which is directly relevant to this paper. The second is to use PDE methods in designing deep neural nets.

The interplay between PDEs and machine learning has lead to many interesting works~\cite{sorteberg2018approximating, weinan2018deep,ruthotto2019deep,chen2018neural}. On the one hand, classical numerical schemes for PDEs have guided and provided more interpretability for the success of certain neural network architectures. For example, several works have formulated ResNets as a discretization of a certain differential equation~\cite{weinan2017proposal,lu2018beyond}, which allowed techniques for improving stability from classical numerical schemes to be applied to deep networks as well~\cite{ruthotto2019deep,chang2017multi}. Chen et al.~\cite{chen2018neural} formalized this connection by connecting depth to the time horizon of a dynamical system. On the other hand, the representational power of deep neural networks (DNN) makes it an excellent candidate for replacing certain components of a PDE solver. For example Weinan et al. used DNNs for general variational problems~\cite{weinan2018deep} as well as for high-dimensional parabolic PDEs~\cite{han2018solving}. Such an approach has also been successfully applied to time-dependent PDEs, such as wave propagation~\cite{sorteberg2018approximating}, weather forecasting~\cite{rodrigues2018deepdownscale}, as well as generic non-linear evolution PDEs~\cite{long2018pde}.

In contrast to most existing methods which rely on the use of large deep neural
nets, our method uses only a shallow network with about 50K parameters, making
it an extremely efficient solution while maintaining a surprisingly high
fidelity. This is made possible by the use of BLADE, an efficient trainable and
adaptive filtering framework that is applicable to a wide range of imaging
applications.

We are interested particularly in time-dependent PDEs appearing in image
processing, such as total variation (TV) flow,
\begin{equation}
\left\{\begin{array}{l}
\partial_t u = \operatorname{div}\bigl(
\frac{\nabla u}{|\nabla u|}\bigr), \\
u(x, y, 0) = f(x, y).
\end{array}\right.
\end{equation}
Above, $u(x, y, t)$ is the image, $(x, y)$ are the spatial coordinates, and $t$
is the evolution time. Such PDEs appear routinely in variational methods as the
gradient flow for the minimization of some
objective~\cite{andreu2001minimizing}.

Consider an equation of the form
\begin{equation}\label{e:hyperbolic-pde}
\partial_t u = F(u, \partial_x u, \partial_y u, \ldots),
\end{equation}
where $F$ is a function of $u$, its gradient, and possibly higher spatial
derivatives. Let $\vv{u}$ denote the image in discrete space with pixels
$u_{m,n}$. A straightforward conventional approach to solving
(\ref{e:hyperbolic-pde}) is to use finite differences to approximate spatial
derivatives and apply explicit Euler to integrate in time,
\begin{equation}
\vv{u}^{(k+1)} = \vv{u}^{(k)} + \dt F(\vv{u}^{(k)},
D_x \vv{u}^{(k)}, D_y \vv{u}^{(k)}, \ldots).
\end{equation}
The finite differences ($D_x, D_y$) are usually chosen with a localized
stencil, e.g.\ the forward finite difference with a $2\times 1$ footprint:
\begin{equation}
(D_x \vv{u})_{m,n} = \tfrac{1}{\dx}(
u_{m+1,n} - u_{m,n}) = \partial_x u + O(\dx).
\end{equation}
Higher-order-accurate finite differences have limited use in image processing as
their use assumes a several-times continuously differentiable function
(required to expand higher terms of the Taylor series). However, natural
images have relatively low regularity as they contain edges and texture. As a result, low-order,
localized differences are most appropriate and common practice.

On the other hand, this means that each pixel of $\vv{u}^{(k+1)}$
has a small, localized domain of dependence in $\vv{u}^{(k)}$. So information
propagates slowly, and a small $\dt$ is necessary for reliable solution
(the Courant--Friedrichs--Lewy condition~\cite{evans2010}).

Additionally, although numerically convenient, it is well known that explicit
Euler becomes unstable if $\dt$ is too large. 
This
problem tends to be more severe if $F$ includes higher-order spatial derivatives.

We train BLADE to approximate $M$ timesteps of a reference solution. To get high-quality training data, we apply a reference method, then subsample both temporally and spatially. With this approach, BLADE is trained to capture longer time dependency, addressing the small timestep issue. Additionally, since the training data is created at higher spatial resolution and then coarsened,
BLADE is trained to super-resolve spatial derivatives in the equation.

\begin{figure}[!h]
\centering
\mbox{\beginpgfgraphicnamed{images/blade-basic-net}%

\begin{tikzpicture}[>=latex,scale=1.4]

\coordinate [xshift=4pt] (boxtl) at (-3.4,0.85);
\coordinate [xshift=-4pt] (boxtr) at (3.4,0.85);

\draw [draw=orange!75] (-0.7,1.25) -- (boxtl)
  (0.7,1.25) -- (boxtr);

\fill [rounded corners=4pt,draw=orange!75,fill=orange!15]
(-3.4,-1.65) rectangle (3.4,0.85);

\node (Blade) at (0,1.5) [draw=black,fill=PlotColorB!50,rectangle,rounded corners=2pt,inner sep=5pt] {\small$\operatorname{BLADE}$};
\draw [<-] (Blade) -- +(-3em,0) node [left] {\small $\vphantom{x}\smash{\vv{z}}$};
\draw [->] (Blade) -- +(3em,0) node [right] {\small $\vphantom{x}\smash{\vv{\Hat{u}}}$}; 

\begin{scope}[xshift=1.2mm,yshift=-1mm]
\node (selector) at (-0.9,-0.8) [draw=black,fill=gray!30,rectangle,rounded corners=2pt] {\small $s(i)$};
\node (switch) at (-0.25,0) [fill, circle,inner sep=0.9pt] {};
\coordinate (join) at (1.4, 0);

\draw (switch) -- +(-5.2em,0) node [left, xshift=0.8em, yshift=-1ex, align=center,text width=5em,inner sep=0pt] 
{\small $\vv{R}_i \vv{z}$ \\[-2pt] \footnotesize input patch};
\draw [rounded corners=2pt] (-1.65,0) -- (-1.65,-0.8) [->] -- (selector);
\draw [rounded corners=2pt] (selector) -- (-0.25,-0.8) [->] -- (switch);

\node (h0) at (0.85,0.48) [draw=black,fill=PlotColorA!40,rectangle,rounded corners=2pt] {\small$\vv{h}^0$};
\node (h1) [yshift=-0.6cm] at (0.85,0.48) [draw=black,fill=PlotColorA!40,rectangle,rounded corners=2pt] {\small$\vv{h}^1$};
\node (h2) [yshift=-1.2cm] at (0.85,0.48) [draw=black,fill=PlotColorA!40,rectangle,rounded corners=2pt] {\small$\vv{h}^2$};

\foreach \k in {-1, 0, 1} {
  \fill [yshift=-1.8cm,xshift=0.85cm,yshift=0.54cm,scale=0.14] (0,\k cm) circle (4.8pt);
}
\draw (join) -- +(0,-9mm);
\draw [dash pattern=on 1.7pt off 1.3pt] (join) ++(0,-9mm) -- +(0,-12pt);

\draw [thick,gray,xshift=-0.25cm] (-35:5mm) arc (-35:35:5mm);
\draw [shorten >=3.5pt] (switch) -- (h0);

\draw [rounded corners=2pt] (h0) -- (h0-|join) -- (join);
\draw [rounded corners=2pt] (h1) -- (h1-|join) -- (join);
\draw [rounded corners=2pt] (h2) -- (h2-|join) -- (join);
\draw [->] (join) -- (2,0) node [right,xshift=-0.85em,yshift=-1ex,align=center, text width=4em, inner sep=0pt] 
{\small $\Hat{\vv{u}}_i$ \\[-2pt] \footnotesize output pixel};
\end{scope}

\end{tikzpicture}%
\endpgfgraphicnamed}
\caption{BLADE, a shallow 2-layer structure, taking image $\vv{z}$ as input and
producing image $\Hat{\vv{u}}$ as output.}
\end{figure}
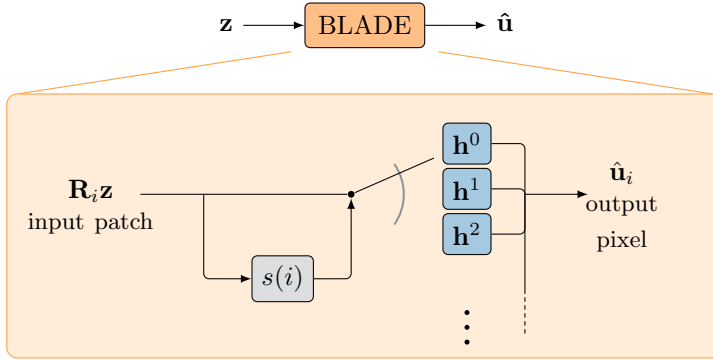

\subsection{Notation}

We denote the PDE solution by $u(x, y, t)$, a function of continuous space and
time, where $(x, y) \in \Omega \subset \mathbb{R}^2$ is the spatial domain
(usually a bounded rectangle). We follow the conventional notations
$\partial_t u$ for partial derivative of $u$ with respect to $t$, $\nabla$ for
spatial gradient, $\div$ for spatial divergence, and $\Delta := \div\nabla$
for spatial Laplacian.

In numerical implementation, we consider an $M\times N$ discrete image as a vector
$\vv{f}\in\mathbb{R}^{M\cdot N}$. Subscripts $f_{m,n}$ or $f_i$,
$i = (m,n) \in \mathbb{Z}^2$, denote the sample at $x = m \dx$, $y = n \dx$.
We denote by $\vv{u}^{(k)}$ the discrete solution at time $t = k\dt$.

\section{BLADE}

Best linear adaptive enhancement (BLADE) is a trainable adaptive filtering
framework that is simple, fast, and useful for a wide range of imaging
problems. It is the generalization of the Rapid and Accurate Image
Super-Resolution (RAISR) method~\cite{romano2017raisr} to tasks other than
image upscaling.

The BLADE network takes an input image $\vv{z}$ as input and computes an output
image $\Hat{\vv{u}}$. The network has a set of learnable, locally linear filters with small footprint:
$\vv{h}^0, \vv{h}^1, \ldots$ and a filter selection mechanism $s(i)$
that decides for each output pixel which filter to apply. The $i$th output pixel
is computed by applying filter $\vv{h}^{s(i)}$:
\begin{equation}\label{e:blade-inference}
\Hat{u}_i = \sum_{j\in F} h_j^{s(i)} z_{i+j},
\end{equation}
where $F\subset \mathbb{Z}^2$ denote the footprint of the filter. In
experiments, we will set $F$ to a $5\times 5$ filter footprint. To filter pixels near the image
boundaries, We extend $\vv{z}$ by replicating border pixels.
BLADE may be
seen as a type of spatially-varying convolution. With a spatially constant
selection $s(i) \equiv s$, (\ref{e:blade-inference}) is the
cross-correlation of $\vv{h}^s$ with $\vv{z}$.

Another way to express (\ref{e:blade-inference}) is as extracting an input
patch and computing its dot product with (only) one selected filter. Denoting
patch extraction by $(\vv{R}_i \vv{z})_j = z_{i+j}$, $j\in F$, the $i$th output
pixel is
\begin{equation}
\Hat{u}_i = (\vv{h}^{s(i)})^T \vv{R}_i \vv{z}.
\end{equation}

BLADE can be seen as a shallow two-layer structure, where the first layer selects
the filter, and the second layer applies the filter.
Notably, only the selected filter at each pixel is evaluated. Therefore, in
contrast to conventional convolutional layers, BLADE's inference computation cost is
independent of the number of filters. BLADE is successfully used in
consumer products where the number of filters is in the hundreds to
low thousands~\cite{romano2017raisr,getreuer2018blade,choi2018fast}.

\subsection{Filter selection}
\label{sec:filter-selection}

\begin{figure}[t]
\centering
\begin{tabular}{@{}c@{\,}c@{\,}c@{\,}c@{}}
\small Input & \small Orientation & \small Strength & \small Coherence \\
\includegraphics[width=0.2\textwidth]{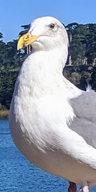} &
\includegraphics[width=0.2\textwidth]{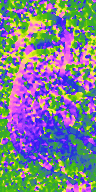} &
\includegraphics[width=0.2\textwidth]{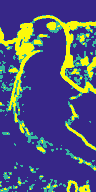} &
\includegraphics[width=0.2\textwidth]{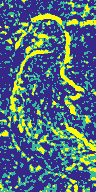}
\end{tabular}
\caption{\label{fig:selection-example}Example of quantized structure tensor features used for filter selection, distinguishing 24 orientations, 3 strength values, and 3 coherence values.}
\end{figure}

An effective choice for the selection mechanism $s(i)$ is to use features
of the $2\times 2$ image structure tensor. Define the structure tensor
\begin{equation}
J(\nabla u) = \left(\begin{array}{cc}
(\partial_x u)^2 & (\partial_x u) \, (\partial_y u) \\
(\partial_x u) \, (\partial_y u) & (\partial_y u)^2
\end{array}\right),
\end{equation}
which is a symmetric $2\times 2$ matrix at each pixel. The tensor is smoothed
with a Gaussian $G_\rho$ of standard deviation $\rho$,
\begin{equation}
J_\rho(\nabla u) = G_\rho * J(\nabla u),
\end{equation}
where convolution is applied spatially to each component. At each pixel, we
compute the eigenvalues $\lambda_1 \ge
\lambda_2$ and dominant eigenvector $\vv{w}_1$ of $J_\rho(\nabla u)$ and use
them to define three features:
\begin{itemize}
\item \textbf{orientation} $= \angle \vv{w}_1$, gradient orientation,
\item \textbf{strength} $= \sqrt{\lambda_1}$, gradient magnitude,
\item \textbf{coherence} $= \frac{\sqrt{\lambda_1} - \sqrt{\lambda_2}}
{\sqrt{\lambda_1} + \sqrt{\lambda_2}}$, local anisotropy.
\end{itemize}
Theses features are quantized to a small number of possible values.
Figure~\ref{fig:selection-example} shows these three quantized features for an
example image. We then consider the quantized values as a three-dimensional
index into the bank of filters. A typical configuration is 24 possible
orientations, 3 strengths, and 3 coherences for a total of $24\cdot 3\cdot 3 =
216$ filters, and flattened to a filter index $\{0,\ldots,215\}$ as
\begin{equation}
s(i) = \mathit{ori}(i) + 24\cdot\bigl(
\mathit{str}(i) + 3 \cdot \mathit{coh}(i)\bigr).
\end{equation}

These structure tensor features enable BLADE to
perform robust edge-adaptive filtering. Other features can be used for
selection. When approximating the Cahn--Hilliard equation for instance,
we will use the input intensity as an additional selection feature.

\subsection{BLADE for PDEs}

We now develop how BLADE can be applied to the solution of hyperbolic PDEs, for
instance the form (\ref{e:hyperbolic-pde}),
\begin{equation*}
\partial_t u = F(u, \partial_x u, \partial_y u, \ldots).
\end{equation*}

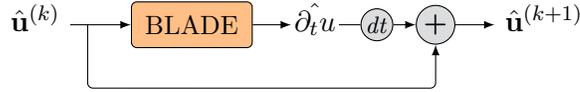
\begin{figure}[!h]
\centering
\mbox{\beginpgfgraphicnamed{images/blade-dt-model}%

\begin{tikzpicture}[>=latex,scale=1.2]

\node (Blade) at (0,0) [draw=black,fill=PlotColorB!50,rectangle,rounded corners=2pt,inner sep=5.2pt] {\small$\operatorname{BLADE}$};
\coordinate (below) at (0,-0.7);
\node (dudt) at (1.1,0) [right, inner sep=1pt] {$\smash{\Hat{\partial_t u}}\vphantom{x}$};
\node (dtscale) at (2.05,0) [draw=black,fill=gray!30,circle,inner sep=0.5pt] {\footnotesize $\dt$};
\node (adder) at (2.7,0) [draw=black,fill=gray!30,circle,inner sep=1pt] {\small\bf +};

\draw [<-] (Blade) -- +(-3.5em,0) node [left] {$\smash{\Hat{\vv{u}}^{(k)}}\vphantom{x}$};
\draw [->] (Blade) -- (dudt);
\draw [->] (dudt) -- (dtscale) -- (adder);
\draw [->] (adder) -- +(1.7em,0) node [right] {$\smash{\Hat{\vv{u}}^{(k+1)}}\vphantom{x}$};

\draw [rounded corners=2pt] (-3em,0) -- +(0,-0.7) -- (below);
\draw [->,rounded corners=2pt] (below) -- (below-|adder) -- (adder);

\end{tikzpicture}%
\endpgfgraphicnamed}
\caption{\label{fig:blade-dt-model}BLADE sequence model based on explicit Euler time integration.}
\end{figure}

Our approach is to estimate the time derivative $\partial_t u$
from $\vv{u}$; that is, $\operatorname{BLADE}(\vv{u}) = \Hat{\partial_t u}$,
where $\operatorname{BLADE}(\vv{u})$ denotes the application of BLADE to
$\vv{u}$ as in equation (\ref{e:blade-inference}) and $\Hat{\cdot}$ denotes that the quantity is an estimate.
We then use explicit Euler integration to advance to the next time step,
\begin{equation}\label{e:blade-dt-model}
\Hat{\vv{u}}^{(k+1)} = \Hat{\vv{u}}^{(k)}
+ \dt \, \operatorname{BLADE}(\Hat{\vv{u}}^{(k)}).
\end{equation}
The above equation is initialized with $\Hat{\vv{u}}^{(0)} = \vv{u}^{(0)}$ and
applied repeatedly to predict the evolution $\vv{u}^{(1)}, \vv{u}^{(2)}, \ldots$ over multiple time steps.

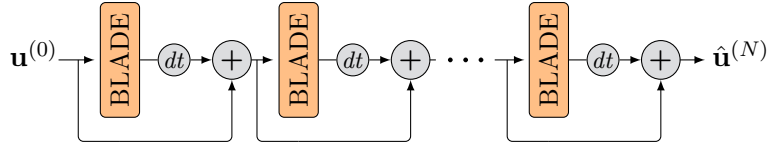
\begin{figure}[!h]
\centering
\mbox{\beginpgfgraphicnamed{images/blade-multiple-steps-net}%

\begin{tikzpicture}[>=latex,scale=1.2]

\draw [->] (-0.68,0) node [left,inner sep=0pt] {$\smash{\vv{u}^{(0)}}\vphantom{x}$} -- +(4mm,0);

\foreach \x in {0, 1, 2} {
  \fill [xshift=1.98*1cm,xshift=1.1cm,xshift=5.8mm] (0.17*\x cm,0) circle (0.75pt);
}
\draw [->,xshift=1.98*2.4cm] (-0.6,0) -- +(3.3mm,0);

\draw [->] (1.2,0) -- +(5mm,0);
\draw [xshift=1.98cm] (1.17,0) -- +(3.5mm,0);
\draw [->,xshift=2.4*1.98cm] (1.2,0) -- +(5.3mm,0) node [right,inner sep=2.5pt]
{$\smash{\Hat{\vv{u}}^{(N)}}\vphantom{x}$};

\foreach \x in {0, 1, 2.4} {
  \begin{scope}[xshift=1.98*\x cm]
  \node (Blade) at (0,0) [draw=black,fill=PlotColorB!50,rectangle,rounded corners=2pt,inner sep=4pt,rotate=90] {\small$\operatorname{BLADE}$};
  \coordinate (below) at (0,-0.9);
  \node (dtscale) at (0.6,0) [draw=black,fill=gray!30,circle,inner sep=0.5pt] {\footnotesize $\dt$};
  \node (adder) at (1.24,0) [draw=black,fill=gray!30,circle,inner sep=1pt] {\small\bf +};

  \draw (Blade) -- (dtscale);
  \draw [->] (dtscale) -- (adder);

  \draw [rounded corners=2pt] (-1.2em,0) -- +(0,-0.9) -- (below);
  \draw [->,rounded corners=2pt] (below) -- (below-|adder) -- (adder);
  \end{scope}
}

\end{tikzpicture}%
\endpgfgraphicnamed}
\caption{\label{fig:blade-multiple-steps-net}BLADE estimation of a stopping time $T = N\dt$ viewed as an $N$-layer
deep network with weights shared across layers.}
\end{figure}

The structure of (\ref{e:blade-dt-model}) may be seen as a spatially-varying
convolutional layer with a skip connection, forming what is called a ``residual'' block in the learning literature~\cite{he2016deep} (Figure~\ref{fig:blade-dt-model}). Its repeated
application may be seen as a deeper model composed of relatively simple blocks (Figure~\ref{fig:blade-multiple-steps-net}). If we are interested only
in some stopping time $T = N\dt$, the model could be considered as an $N$-layer
deep network with weight sharing across layers that produces $u^{(N)}$ as the
final output.

\begin{figure}[!h]
\centering
\mbox{\beginpgfgraphicnamed{images/blade-dt-model-midpoint}%

\begin{tikzpicture}[>=latex,xscale=0.95,scale=1.2]

\node (Blade) at (0.1,0) [draw=black,fill=PlotColorB!50,rectangle,rounded corners=2pt,inner sep=4pt,rotate=90] {\small$\operatorname{BLADE}$};
\coordinate (below) at (0,-0.9);
\node (dtscale) at (0.75,0) [draw=black,fill=gray!30,circle,inner sep=0.5pt] {\footnotesize $\tfrac{\dt}{2}$};
\node (adder) at (1.5,0) [draw=black,fill=gray!30,circle,inner sep=1pt] {\small\bf +};

\draw [->] (-0.85,0) node [left,inner sep=0pt] {$\smash{\vv{u}^{(k)}}\vphantom{x}$} -- (Blade);

\draw (Blade) -- (dtscale);
\draw [->] (dtscale) -- (adder);
     
\draw [rounded corners=2pt] (-0.4,0) -- +(0,-0.9) -- (below);
\draw [->,rounded corners=2pt] (below) -- (below-|adder) -- (adder);

\node (Blade2) at (2.27,0) [draw=black,fill=PlotColorB!50,rectangle,rounded corners=2pt,inner sep=4pt,rotate=90] {\small$\operatorname{BLADE}$};
\coordinate (below2) at (0,-1.1);
\node (dtscale2) at (2.85,0) [draw=black,fill=gray!30,circle,inner sep=0.5pt] {\footnotesize $\dt$};
\node (adder2) at (3.5,0) [draw=black,fill=gray!30,circle,inner sep=1pt] {\small\bf +};

\draw [->] (adder) -- (Blade2);
\draw (Blade2) -- (dtscale2);
\draw [->] (dtscale2) -- (adder2);

\draw [rounded corners=2pt] (-0.6,0) -- +(0,-1.1) -- (below2);
\draw [->,rounded corners=2pt] (below2) -- (below2-|adder2) -- (adder2);

\draw [->] (adder2) -- +(5.8mm,0) node [right,inner sep=2.5pt]
{$\smash{\Hat{\vv{u}}^{(k+1)}}\vphantom{x}$};

\end{tikzpicture}%
\endpgfgraphicnamed}
\caption{\label{fig:blade-dt-model-midpoint}BLADE with explicit midpoint time integration.}
\end{figure}
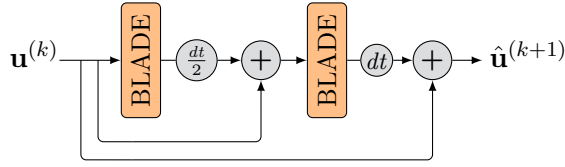

In the time dimension, explicit Euler integration could be substituted with a higher-order method at the cost of more network evaluations per time step. For instance explicit midpoint integration could be done as
\begin{equation}
\begin{aligned}
\Hat{\vv{u}}^{(k+1)} &= \Hat{\vv{u}}^{(k)}
+ \dt \, \operatorname{BLADE}\bigl( \\
&\quad \Hat{\vv{u}}^{(k)} + \tfrac{\dt}{2}
\, \operatorname{BLADE}(\Hat{\vv{u}}^{(k)})\bigr).
\end{aligned}
\end{equation}
This may be seen as a network structure with two skip connections (Figure~\ref{fig:blade-dt-model-midpoint}).

\subsection{BLADE training}

\begin{figure}[!h]
\centering%
\mbox{\beginpgfgraphicnamed{images/target_frame_sequence}%
\input{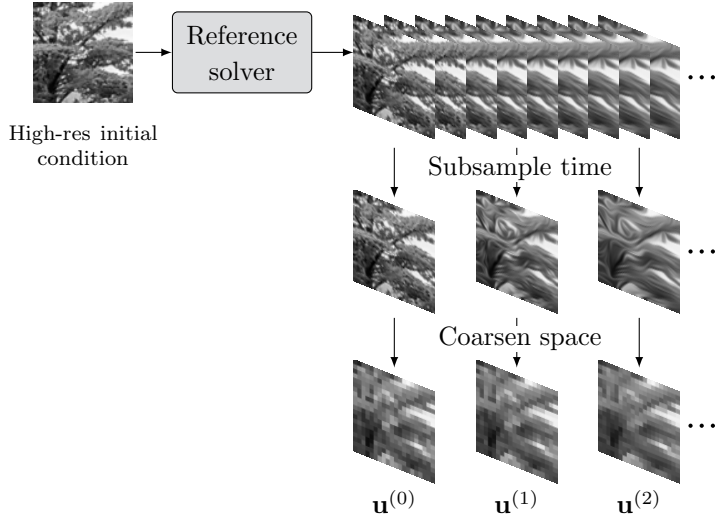}%
\endpgfgraphicnamed}%
\caption{\label{f:target-frame-sequence}Preparing a target frame sequence
for training.}
\end{figure}

Next we describe how we train the BLADE sequence model for approximating the
solution of the described anisotropic diffusion PDEs.
To create high-quality data for training, we perform the following steps, as
illustrated in Figure~\ref{f:target-frame-sequence}:
\begin{enumerate}
\item We begin with an initial condition that is, say, $4\times$ higher spatial
resolution than the operating resolution at which we intend to run BLADE. A
reference implementation of the PDE is
executed on this initial condition with fine time step
$\dt_\mathit{HR}$. This
is such that quantization both spatially and temporally is fine, so that the
computed reference solution is high quality.

\item We subsample temporally, keeping every $M$th
frame, to coarsen time resolution to $\dt = M \dt_\mathit{HR}$.

\item The frames are spatially downscaled using an area-averaging kernel,
reducing space to the target operating resolution. The resulting sequence of
frames is the training target.
\end{enumerate}
We can generate target frame sequences of high quality in this manner, possibly
exceeding the quality of simply running the reference method directly at the
operating resolution.

In training, we unroll ten time steps of (\ref{e:blade-dt-model}).
Training examples are formed by partitioning a target frame
sequence into windows of 11 consecutive frames, where the first frame is the
input to the model, and the later 10 frames $\vv{u}^{(k)}$, $k=1,\ldots,10$,
are compared against model predictions $\Hat{\vv{u}}^{(k)}$ with a summed
squared $L^2$ training loss:
\begin{equation}
L(\Hat{\vv{u}}) = \sum_{k=1}^{10} \|\Hat{\vv{u}}^{(k)} - \vv{u}^{(k)}\|_2^2.
\end{equation}

We remark that instead of $L^2$, other differentiable metrics could be used as
the loss without impact on cost at inference time. For PDEs on
natural images, mean structural similarity (SSIM)~\cite{wang2004image},
difference in VGG activations~\cite{johnson2016perceptual}, or
neural image assessment (NIMA)~\cite{talebi2018learned} could be used to
optimize for perceptual quality of the approximation.

We use TensorFlow~\cite{abadi2016tensorflow} in this work to train BLADE. The
BLADE filtering formula (\ref{e:blade-inference}) is implemented as an op, a
function taking the filters $(\vv{h}^0, \vv{h}^1, \ldots)$, selection $s$, and
image $\vv{z}$ as inputs and producing a filtered image $\Hat{u}$ as output. We
backpropagate the training loss gradient $\tfrac{\partial L}{\partial
\Hat{\vv{u}}}$ as
\begin{equation}\label{e:blade-gradient}
\left\{
\begin{aligned}
\frac{\partial L}{\partial h^k_j}
&= \sum_{\mathclap{i : s(i) = k}} z_{i + j} \frac{\partial L}{\partial \Hat{u}_i}, \\
\frac{\partial L}{\partial z_i}
&= \sum_{j\in F} h^{s(i - j)}_j \frac{\partial L}{\partial \Hat{u}_{i + j}}.
\end{aligned}\right.
\end{equation}

Note that (\ref{e:blade-inference}) is discontinuous with respect to selection, due to the filter
lookup. Therefore, no $\tfrac{\partial L}{\partial s}$ gradient is backpropagated. This is suboptimal,
since training ignores how selection could be changed (i.e.\ due to changes
in preceding operations) to improve the training objective. Nevertheless, we
find in experiments that training rate and convergence are satisfactory
with gradient equation (\ref{e:blade-gradient}).

\subsection{Properties}

The proposed structure is quite flexible, even after training. It is reasonable
to run the time step prediction (\ref{e:blade-dt-model}) with
a smaller $\dt$ than used in training if it is desired to
estimate $u$ at a specific point in time. Moreover, it is reasonable to add
other terms to the right hand side of
(\ref{e:hyperbolic-pde}) not seen during training, simply by adding them in
the explicit Euler step. We will show for example that BLADE trained to perform
TV flow $\partial_t u = \div(\grad u/|\grad u|)$ can be applied as
\begin{equation}
\Hat{\vv{u}}^{(k+1)} = \vv{u}^{(k)} +
\dt \bigl(\operatorname{BLADE}(\vv{u}^{(t)})
+ \lambda A^T(u^{(0)} - Au)\bigr),
\end{equation}
where $A$ is a blurring operator to approximate
\begin{equation}
\partial_t u = \div\bigl(\tfrac{\grad u}{|\grad u|}\bigr) +
\lambda A^T(f - Au)
\end{equation}
whose steady state solution is TV-regularized deblurring. So once
trained, a BLADE network is potentially useful for multiple applications without
retraining.

We remark on several other attractive properties of this approach:
\begin{itemize}
\item BLADE learns to approximate spatial derivatives in the equation. With high-quality training
examples created at finer resolution, BLADE's approximation is possibly superior to finite differences.

\item Training may help accuracy in the temporal dimension as well, since
the explicit Euler time integration is incorporated in the model and trained end-to-end.

\item Training the network unrolled over multiple time steps encourages BLADE to
be stable, or in the very least to have controlled error growth over the
unrolled steps.
\end{itemize}

\section{Anisotropic diffusion PDEs}

In this section, we investigate using BLADE to approximate PDEs for several image processing tasks: TV flow, Perona--Malik, coherence enhancing diffusion, and Cahn--Hilliard.

\paragraph{Total variation flow}
Total variation (TV) flow is the edge-preserving diffusion
\begin{equation}\label{e:tv_flow_pde}
\partial_t u = \div\Bigl(\frac{\nabla u}{|\nabla u|}\Bigr).
\end{equation}
TV flow arises from gradient descent of the TV seminorm
$\int |\nabla u| \, dx$~\cite{rudin1992nonlinear}. This makes it of
interest in implementing TV-regularized variational methods. We use as reference
implementation the explicit scheme suggested by Rudin, Osher, and
Fatemi~\cite{rudin1992nonlinear}:
\begin{equation}
\begin{aligned}
\lefteqn{u_{m,n}^{(k+1)} = u^{(k)}_{m,n} + \dt\Biggl[} \\
&
D_-^x \Biggl(\frac{D_+^x u_{m,n}^{(k)}}{\sqrt{(D_+^x u_{m,n}^{(k)})^2
+ \min(D_+^y u_{m,n}^{(k)}, D_-^y u_{m,n}^{(k)})^2}}\Biggr) \\
&+ D_-^y \Biggl(\frac{D_+^y u_{m,n}^{(k)}}{\sqrt{(D_+^y u_{m,n}^{(k)})^2
+ \min(D_+^x u_{m,n}^{(k)}, D_-^x u_{m,n}^{(k)})^2}}\Biggr)\Biggr].
\end{aligned}
\end{equation}
where $D_+^x u_{m,n} := u_{m+1,n} - u_{m,n}$ and $D_-^x u_{m,n} := u_{m,n} - u_{m-1,n}$ denote forward and backward finite differences in the x direction, and similarly $D_+^y$ and $D_-^y$ denote finite
differences in the y direction.

\paragraph{Perona--Malik}
The Perona--Malik equation~\cite{perona1990scale} is another edge-preserving diffusion,\footnote{As noted by
Esedoḡlu~\cite{esedoglu2001analysis}, the Perona--Malik equation is often thought of as
$\partial_t u = \div\bigl(g(|\nabla u|) \nabla u\bigr)$, but more precisely,
Perona and Malik's intention and discretization are suggestive of the
anisotropic form written above.}
\begin{equation}\label{e:perona_malik_pde}
\partial_t u = \partial_x \bigl( g(|u_x|^2) u_x \bigr)
+ \partial_y \bigl(g(|u_y|^2) u_y\bigr),
\end{equation}
in which we set $g(s) = 1/(1+s/c^2)$.

We use the original scheme suggested by Perona and
Malik~\cite{perona1990scale} as reference implementation:
\begin{equation}
\begin{aligned}
u_{m,n}^{(k+1)} = u^{(k)}_{m,n} + \frac{\dt}{\dx}\Bigl[
& g(|D_+^x u_{m,n}^{(k)}|^2) D_+^x u_{m,n}^{(k)} \\
{-}\,\, & g(|D_-^x u_{m,n}^{(k)}|^2) D_-^x u_{m,n}^{(k)} \\
{+}\,\, & g(|D_+^y u_{m,n}^{(k)}|^2) D_+^y u_{m,n}^{(k)} \\
{-}\,\, & g(|D_-^y u_{m,n}^{(k)}|^2) D_-^y u_{m,n}^{(k)} \Bigr].
\end{aligned}
\end{equation}

\paragraph{Coherence enhancing diffusion}
Coherence enhancing diffusion (CED) is an anisotropic diffusion introduced by
Weickert~\cite{weickert1998anisotropic},
\begin{equation}\label{e:coherence-enhancing-diffusion}
\partial_t u = \div(D \nabla u),
\end{equation}
where $D$ is a tensor, a $2\times 2$ matrix at each pixel, constructed from the
structure tensor of $u$ as
\begin{align}
\mu_1 &= \alpha, \\
\mu_2 &= \alpha + (1 - \alpha) \exp\bigl(-C / (\lambda_1 - \lambda_2)^2\bigr), \\
D &= \mu_1 \vv{w}_1 \vv{w}_1^T + \mu_2 \vv{w}_2 \vv{w}_2^T,
\end{align}
where $\alpha$ and $C$ are positive parameters, and as described previously in
\S\ref{sec:filter-selection}, $\lambda_1, \lambda_2, \vv{w}_1$, $\vv{w}_2$
are the eigenvalues and eigenvectors of $J_\rho(\nabla u)$. The CED equation has
an oil painting like effect, diffusing the image along oriented features in a
flowing self-reinforcing manner.

We use the scheme developed by Weickert and Scharr~\cite{weickert2002scheme}
as reference implementation, in which spatial derivatives in
(\ref{e:coherence-enhancing-diffusion}) are discretized with the
rotation-optimized $3\times 3$ filters
\begin{equation}
\frac{1}{32}\begin{pmatrix}
-3 & 0 & 3 \\
-10 & 0 & 10 \\
-3 & 0 & 3
\end{pmatrix}, \;
\frac{1}{32}\begin{pmatrix}
3 & 10 & 3 \\
0 & 0 & 0 \\
-3 & -10 & -3
\end{pmatrix}.
\end{equation}

\paragraph{Cahn--Hilliard}
The Cahn--Hilliard equation is
\begin{equation}\label{e:cahn_hilliard_pde}
\partial_t u = \Delta\bigl( W'(u) - \gamma \Delta u \bigr)
\end{equation}
where $\gamma$ is a positive parameter and $W'$ denotes the derivative of the
double-well potential $W(u) = u^2 (u - 1)^2$.

This equation describes the process of phase
separation of a binary fluid. If applied to a grayscale image, the equation
drives the image toward binary 0/1 values (the minima of $W$).
The equation conserves the total mass $\int_\Omega u \, dx$, so regions of
darker graylevels tend to evolve toward islands of white surrounded
by black, and symmetrically for lighter grays, creating a dithering-like
effect.

Bertozzi, Esedoḡlu and Gillette~\cite{bertozzi2007inpainting} showed that the
Cahn--Hilliard equation when applied on
binary images has the effect of bridging small gaps, which makes it useful
for inpainting binary images.

Following Bertozzi et al.~\cite{bertozzi2007inpainting}, we use the following
semi-implicit scheme as reference implementation
\begin{equation}
\vv{u}^{(k+1)} + \dt \gamma \Delta^2 \vv{u}^{(k+1)}
= \vv{u}^{(k)} + \dt \Delta W'(\vv{u}^{(k)}).
\end{equation}
The equation is solved for $\vv{u}^{(k+1)}$ by inverting
$(I + \dt \gamma \Delta^2)$ in the Fourier domain.

\subsection{Conservative Model}
Several of the anisotropic diffusion PDEs in the previous section have the form of a conservation law:
\begin{equation}\label{e:conservative_pde}
\partial_t u = \div G(u),
\end{equation}
where $G$ is
\begin{itemize}
    \item TV flow (\ref{e:tv_flow_pde}): $G(u) = \grad u / |\grad u|$,
    \item Coherence enhancing diffusion (\ref{e:coherence-enhancing-diffusion}):
            $G(u) = D \grad u)$,
    \item Cahn--Hilliard (\ref{e:cahn_hilliard_pde}):
            $G(u) = \grad \bigl(W'(u) - \gamma \Delta u\bigr)$.
\end{itemize}
The form (\ref{e:conservative_pde}) implies that total mass $\int u\, dx$ is conserved.
We can integrate over a control area $A$ and apply
divergence theorem to express the right hand side of (\ref{e:conservative_pde}) as fluxes across its boundary
$S$:
\begin{equation}\label{e:pde-flux-form}
\partial_t \Bigl(\int_A u \, dx\Bigr) = \int_A \div G(u) \, dx
= \oint_S G(u) \cdot \mathbf{n} \, dS.
\end{equation}
Provided no flux crosses the image boundaries, flux exiting one element enters a
neighboring element so that total mass is conserved.

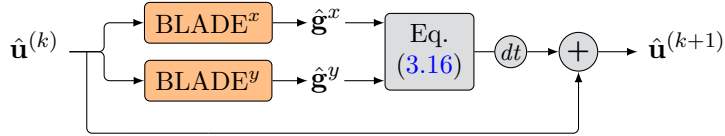
\begin{figure}[!h]
\centering
\mbox{\beginpgfgraphicnamed{images/blade-flux-model}%

\begin{tikzpicture}[>=latex,scale=1.2]

\node (Bladex) at (0,0.32) [draw=black,fill=PlotColorB!50,rectangle,rounded corners=2pt,inner sep=4pt] {\small$\operatorname{BLADE}^x$};
\node (Bladey) at (0,-0.32) [draw=black,fill=PlotColorB!50,rectangle,rounded corners=2pt,inner sep=4pt] {\small$\operatorname{BLADE}^y$};
\coordinate (below) at (0,-0.9);
\node (gx) at (1.1,0.32) [right, inner sep=1pt] {$\smash{\Hat{\vv{g}}^x}\vphantom{x}$};
\node (gy) at (1.1,-0.32) [right, inner sep=1pt] {$\smash{\Hat{\vv{g}}^y}\vphantom{x}$};
\node (fluxeq) at (2.45,0) [draw=black,fill=gray!30,rectangle,rounded corners=2pt,inner sep=4pt,align=center,text width=9mm]
{\small Eq. \\[-2pt] (\ref{e:fluxeq})};

\node (dtscale) at (3.35,0) [draw=black,fill=gray!30,circle,inner sep=0.5pt] {\footnotesize $\dt$};
\node (adder) at (4.1,0) [draw=black,fill=gray!30,circle,inner sep=1pt] {\small\bf +};

\draw [rounded corners=2pt] 
(-4em,0) node [left] {$\smash{\Hat{\vv{u}}^{(k)}}\vphantom{x}$}
-- (-3em,0) -- (-3em,0.32) [->] -- (Bladex);
\draw [->] (Bladex) -- (gx);
\draw [->,shorten >=6mm] (gx) -- (gx-|fluxeq);
\draw [rounded corners=2pt] 
(-4em,0) -- (-3em,0) -- (-3em,-0.32) [->] -- (Bladey);
\draw [->] (Bladey) -- (gy);
\draw [->,shorten >=6mm] (gy) -- (gy-|fluxeq);

\draw [->] (fluxeq) -- (dtscale) -- (adder);
\draw [->] (adder) -- +(1.7em,0) node [right] {$\smash{\Hat{\vv{u}}^{(k+1)}}\vphantom{x}$};

\draw [rounded corners=2pt] (-3.5em,0) -- +(0,-0.9) -- (below);
\draw [->,rounded corners=2pt] (below) -- (below-|adder) -- (adder);

\end{tikzpicture}%
\endpgfgraphicnamed}
\caption{BLADE flux model for conservative laws.}
\end{figure}

\begin{figure}[h]
\centering
\includegraphics[width=0.4\textwidth]{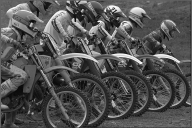}
\includegraphics[width=0.4\textwidth]{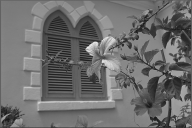}
\caption{\label{f:kodak7-input} Input images to BLADE in
  Fig.~\ref{f:image-pde-table}, $4\times$ coarsened
grayscale versions of Kodak images \#5 and \#7.}
\end{figure}

\begin{figure*}[t]
\centering
\begin{tabular}{@{}cc@{\,}c@{\,}c@{\,}c@{}}
& \small TV flow & \small Perona--Malik & \small CED & \small Cahn--Hilliard \\
\raisebox{6mm}{\small Reference} & \includegraphics[width=0.21\textwidth]{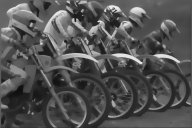} &
\includegraphics[width=0.21\textwidth]{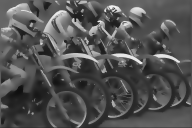} &
\includegraphics[width=0.21\textwidth]{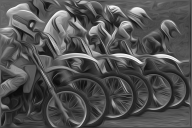} &
\includegraphics[width=0.21\textwidth]{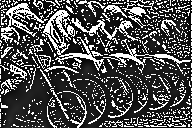} \\
\raisebox{6mm}{\small BLADE} & \includegraphics[width=0.21\textwidth]{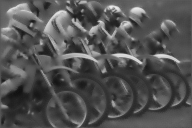} &
\includegraphics[width=0.21\textwidth]{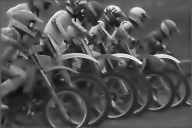} &
\includegraphics[width=0.21\textwidth]{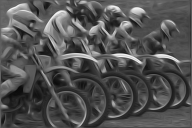} &
\includegraphics[width=0.21\textwidth]{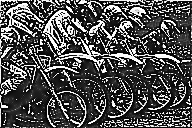}\\

\raisebox{6mm}{\small Reference} & \includegraphics[width=0.21\textwidth]{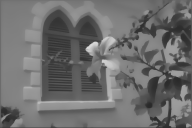} &
\includegraphics[width=0.21\textwidth]{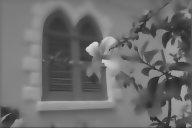} &
\includegraphics[width=0.21\textwidth]{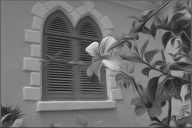} &
\includegraphics[width=0.21\textwidth]{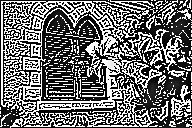} \\
\raisebox{6mm}{\small BLADE} & \includegraphics[width=0.21\textwidth]{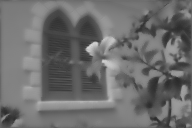} &
\includegraphics[width=0.21\textwidth]{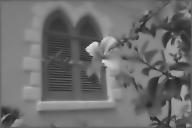} &
\includegraphics[width=0.21\textwidth]{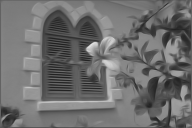} &
\includegraphics[width=0.21\textwidth]{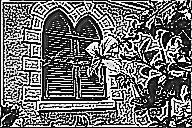}
\end{tabular}
\caption{\label{f:image-pde-kodak-07} First and third rows: Results computed by reference implementation at high resolution.
Second and fourth rows: Approximations with BLADE. The input image to BLADE is shown in Fig.~\ref{f:kodak7-input}.}
\end{figure*}

\begin{figure*}[t]
\centering
\begin{tabular}{@{}c@{\,}c@{\,}c@{\,}c@{}}
\small Color input & \small BLADE TV flow & \small BLADE Perona--Malik & \small BLADE CED \\
\includegraphics[width=0.22\textwidth]{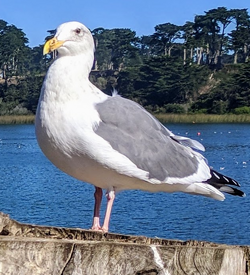} &
\includegraphics[width=0.22\textwidth]{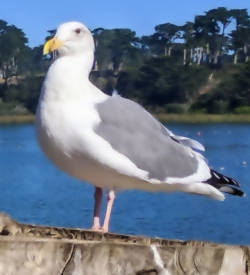} &
\includegraphics[width=0.22\textwidth]{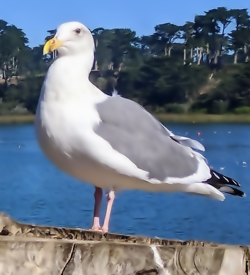} &
\includegraphics[width=0.22\textwidth]{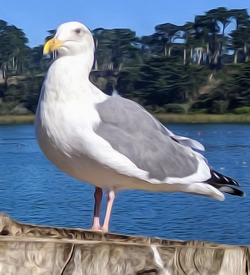}
\end{tabular}
\caption{\label{f:image-pde-table} BLADE PDE approximations applied to a color
input image.
}
\end{figure*}

\begin{figure}[h]
\centering
\includegraphics[width=0.8\textwidth]{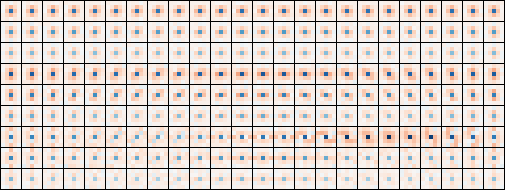}
\caption{\label{f:filters-tv-flow} BLADE filters for TV flow. [Different orientation bins along columns,
and strength and then coherence along rows. There are 24 orientations, 3 strengths, and 3 coherences,
and $\rho = 1.0$.]}
\end{figure}

\begin{figure}[h]
\centering
\includegraphics[width=0.8\textwidth]{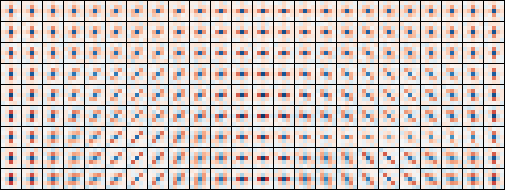}
\caption{\label{f:filters-ced} BLADE filters for CED.}
\end{figure}

\begin{figure}[h]
\centering
\includegraphics[width=0.8\textwidth]{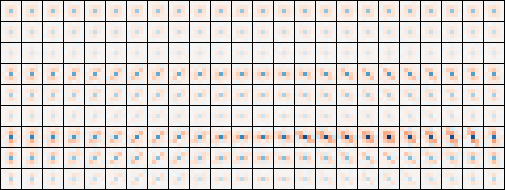}
\caption{\label{f:filters-perona-malik} BLADE filters for Perona--Malik.}
\end{figure}

\begin{figure}[h]
\centering
\includegraphics[width=0.8\textwidth]{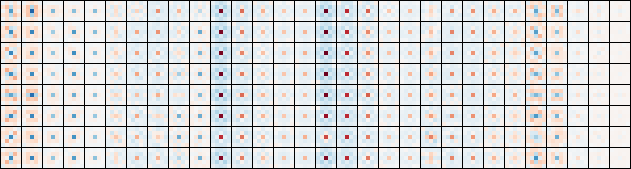}
\caption{\label{f:filters-cahn-hilliard} BLADE filters for Cahn--Hilliard. Different orientation bins along rows, and strength and then intensity along columns. There are 8 orientations, 5 strengths, and 6 intensities.}
\end{figure}

Finite volume methods partition the domain into volume elements and
implement the right hand side of (\ref{e:pde-flux-form}) as a sum of fluxes
between adjacent elements, ensuring that the discrete scheme conserves mass.
We can follow this approach to make a BLADE-based finite volume method that
conserves mass. Let $\operatorname{BLADE}^x$ and $\operatorname{BLADE}^y$ denote
two BLADE networks (with independent filters and selection rule) that estimate
respectively $G(u)^x_{m+1/2,n}$ and $G(u)^y_{m,n+1/2}$ (the $x$ and $y$
components of $G(u)$ at boundary midpoints),
\begin{align}
\operatorname{BLADE}^x(u)_{m,n} &= \Hat{G}^x_{m+1/2,n} \approx G(u)^x_{m+1/2,n}, \\
\operatorname{BLADE}^y(u)_{m,n} &= \Hat{G}^y_{m,n+1/2} \approx G(u)^y_{m,n+1/2}.
\end{align}
Fluxes across the image boundaries are set to zero (Neumann boundary
condition). The flux estimates $\Hat{G}^x$ and $\Hat{G}^y$ are then summed as
follows to estimate $\partial_t u$:
\begin{equation}\label{e:fluxeq}
\begin{aligned}
\Hat{\partial_t u}_{m,n} = &\Hat{G}^x_{m+1/2,n} - \Hat{G}^x_{m-1/2,n} \\
&{+}\,\, \Hat{G}^y_{m,n+1/2} - \Hat{G}^y_{m,n-1/2}.
\end{aligned}
\end{equation}
The BLADE model with this time derivative conserves mass.

\subsection{Results}

With all above PDEs, we set parameters and a large enough stopping time
to produce a moderate effect, and approximate them with BLADE model
(\ref{e:blade-dt-model}) with ten time steps.

\subsection{Evaluation}
We compare the image at the stopping time between
the reference implementation and the BLADE approximation.
Figure~\ref{f:image-pde-kodak-07} shows reference vs.\ BLADE for Kodak
images~\#5 and \#7.

The following table lists average PSNR and SSIM over the Kodak Image
Suite (higher is better).

\begin{tabular}{lrr}
& Average PNSR & Average SSIM \\
TV flow & 33.43 & 0.9553 \\
Perona--Malik & 35.68 & 0.9821 \\
CED & 35.54 & 0.9663 \\
Cahn--Hilliard & 12.92 & 0.7737 \\[2ex]
\end{tabular}

The BLADE Cahn--Hilliard approximation is quantitatively poor, yet it is
visually close (Fig.~\ref{f:image-pde-kodak-07}).

\subsection{Color Images}

Even though we train BLADE on grayscale images, the resulting model extends
readily to color images by two minor modifications: first, selection is
performed on the input's luma channel\footnote{Or as described in
\cite{choi2018fast}, a more robust structure tensor analysis can be done
using jointly all color channels, with only a small increase in computation cost.},
and second, the BLADE filtering equation
(\ref{e:blade-inference}) is performed independently on the R, G, B channels.

\section{Combining with Other Terms}

\begin{figure}[ht]
\centering
\begin{tabular}{@{}c@{\,}c@{\,}c@{}}
\small Ground Truth & \small Input & \small Deconvolution \\
\includegraphics[width=0.27\textwidth]{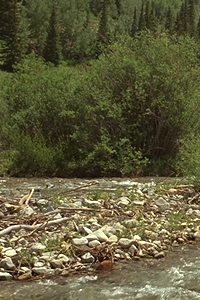} &
\includegraphics[width=0.27\textwidth]{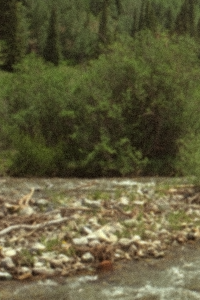} &
\includegraphics[width=0.27\textwidth]{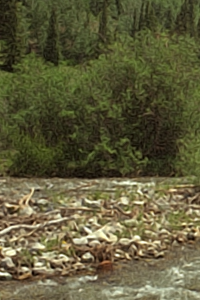}
\end{tabular}
\caption{\label{f:deconv-blade-ced}BLADE CED-regularized deconvolution with
(\ref{e:blade-ced-deconv}). Input has PSNR 23.04~dB and SSIM 0.6424.
The deconvolved image has PSNR 24.66~dB and SSIM 0.7436.}
\end{figure}

\begin{figure}[ht]
\centering
\begin{tabular}{@{}c@{\,}c@{\,}c@{}}
\small Input & \small Upscaled $4\times$ \\
\includegraphics[width=0.35\textwidth]{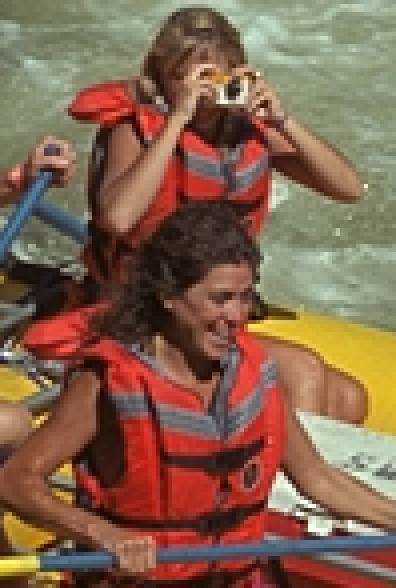} &
\includegraphics[width=0.35\textwidth]{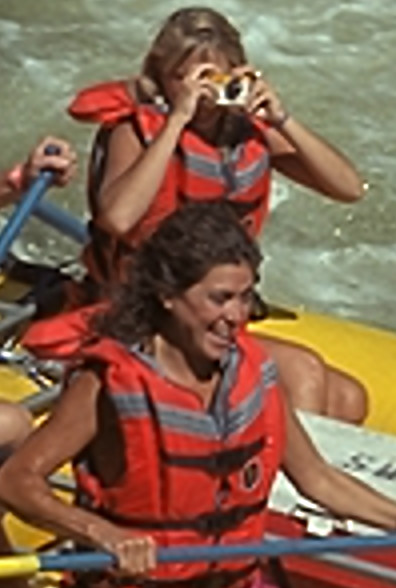}
\end{tabular}
\caption{\label{f:upscale4x-blade-tvf}
BLADE TV-regularized upscaling by factor 4.}
\end{figure}

TV flow, Perona--Malik, and CED are all denoising or regularizing processes in
the sense that they tend to remove noise while preserving image content.
They may be used as regularizers to solve deblurring, inpainting, image
upscaling, and other inverse problems by adding a $+ \lambda A^T(f - Au)$ term
to the PDE.

Consider generically a degradation (observation)
model of the form $f = A u + \mathit{noise}$ where $A$ is a
linear operator and the noise is white Gaussian, then restoration of $u$
for instance with CED is
\begin{equation}\label{e:ced-restoration}
\partial_t u = \div(D \nabla u) + \lambda A^T(f - Au),
\end{equation}
in which $A^T$ denotes the transpose (adjoint) of $A$ and $\lambda$ is a
parameter balancing between matching the degradation model and regularization.

Using the BLADE CED approximation from the previous section, we can implement
CED-regularized restoration (\ref{e:ced-restoration}), without needing to retrain BLADE, as
\begin{equation}\label{e:blade-ced-restoration}
\Hat{\vv{u}}^{(k+1)} = \Hat{\vv{u}}^{(k)} +
\dt \bigl(\operatorname{BLADE}(\Hat{\vv{u}}^{(k)})
+ \lambda A^T(\vv{f} - A\Hat{\vv{u}}^{(k)})\bigr)
\end{equation}
initialized with $\Hat{\vv{u}}^{(0)} = \vv{f}$. BLADE TV flow and BLADE
Perona--Malik approximations can be applied similarly.

\subsection{Nonblind deconvolution}

In the case of nonblind deconvolution with blur kernel $\varphi$, we have
$Au = \varphi * u$ and
\begin{align}
\lambda A^T(\vv{f} - A\Hat{\vv{u}})
&= \lambda \Tilde{\varphi} * (\vv{f} - \varphi * \Hat{\vv{u}}) \nonumber \\
&= -\lambda \Tilde{\varphi} * \varphi * \Hat{\vv{u}}
+ \lambda \Tilde{\varphi} * \vv{f},
\end{align}
where $\Tilde{\varphi}$ denotes spatial reversal of $\varphi$.
The $-\lambda \Tilde{\varphi} * \varphi * \Hat{\vv{u}} $ term can be absorbed
into the BLADE filters, still without needing
to retrain. Let $\vv{h}^k$ denote the BLADE CED filters, then we create
BLADE CED deconvolution filters
\begin{equation}
\vv{h}^k_\mathit{deconv} = \vv{h}^k - \lambda \Tilde{\varphi} * \varphi,
\end{equation}
and (\ref{e:blade-ced-restoration}) with these filters becomes
\begin{equation}\label{e:blade-ced-deconv}
\Hat{\vv{u}}^{(k+1)} = \Hat{\vv{u}}^{(k)} +
\dt \bigl(\operatorname{BLADE}(\Hat{\vv{u}}^{(k)})
+ \lambda \Tilde{\varphi} * \vv{f}\bigr).
\end{equation}
The parameter $\lambda$ balances between deconvolution vs.\ denoising strength.
In Fig.~\ref{f:deconv-blade-ced}, we blurred the input image with
a Gaussian with standard deviation of 1~pixel and added noise of standard
deviation 5. The CED-regularized deconvolution is computed with ten time steps
(\ref{e:blade-ced-deconv}) and $\lambda = 2$.

\subsection{Image upscaling}

For image upscaling, we consider the degradation model
$f = {\downarrow} (\varphi * u) + \mathit{noise}$, where $\varphi$ is the point
spread function and $\downarrow$ denotes subsampling. We perform upscaling as
\begin{equation}\label{e:blade-upscaling}
\begin{array}{@{}r@{}l@{}}
\Hat{\vv{u}}^{(k+1)} = \Hat{\vv{u}}^{(k)} +
\dt \bigl[&\operatorname{BLADE}(\Hat{\vv{u}}^{(k)}) \\
&+ \lambda \Tilde{\varphi} * {\uparrow} \bigl(\vv{f} - {\downarrow} (
\varphi * \Hat{\vv{u}}^{(k)})\bigr)\bigr]
\end{array}
\end{equation}
where $\uparrow$ denotes the transpose operation of $\downarrow$, which is
upsampling by inserting zeros, and $\Hat{\vv{u}}^{(0)}$ is initialized with Lanczos
interpolation of $\vv{f}$.
Figure~\ref{f:upscale4x-blade-tvf} shows factor 4 upscaling of a crop of
Kodak image~\#14 using in this example BLADE TV flow as the regularizer and ten
time steps. The point spread function $\varphi$ is a Gaussian with standard
deviation $0.4$ high-res pixels, and $\lambda = 0.35$.

\begin{figure}[t]
\centering
\begin{tabular}{@{}c@{\,}c@{\,}c@{}}
\small Image & \small Initialization \\
\includegraphics[width=0.38\textwidth]{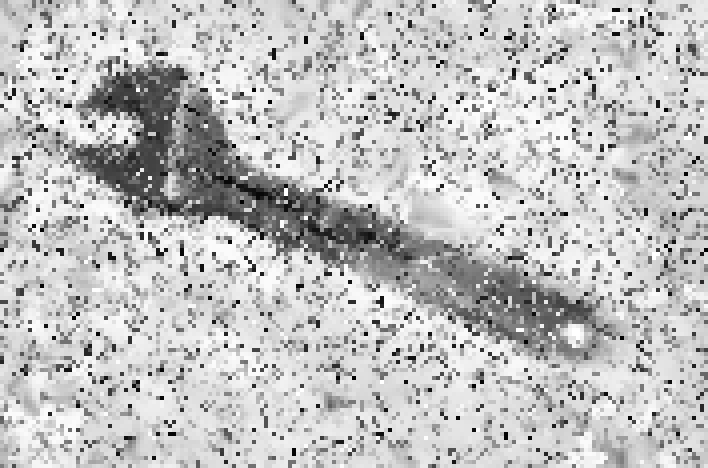} &
\includegraphics[width=0.38\textwidth]{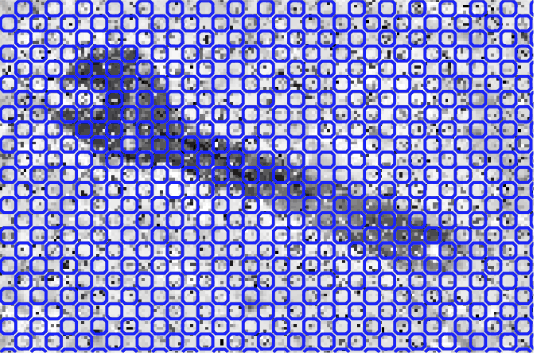} \\
\small Reference implementation & \small BLADE implementation \\
\includegraphics[width=0.38\textwidth]{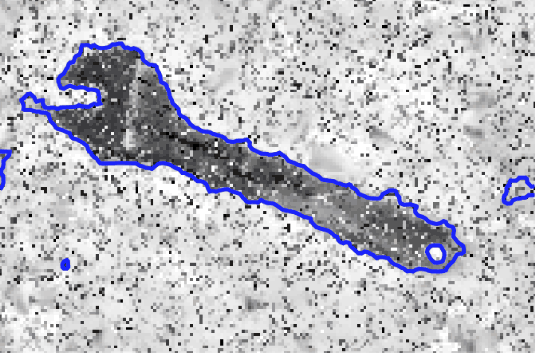} &
\includegraphics[width=0.38\textwidth]{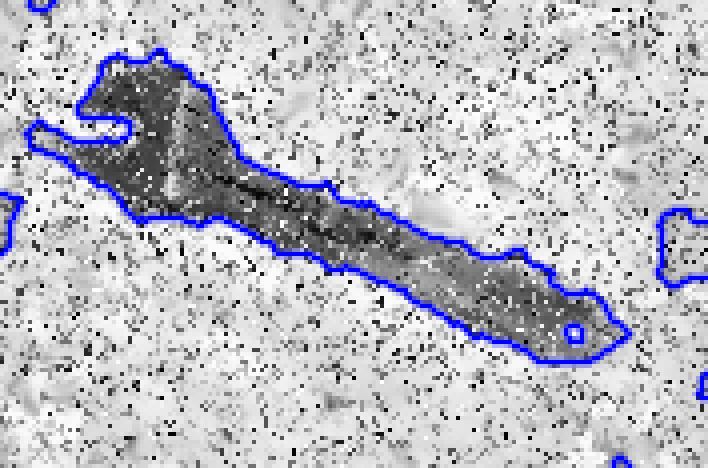}
\end{tabular}
\caption{\label{f:cv-wrench}Example Chan--Vese segmentation using BLADE TV approximation. Top left: input image.
Top right: initialization. Bottom left: segmentation with reference implementation, $\mu = 0.2$. Bottom right:
segmentation with BLADE TV approximation (\ref{e:blade-cv}), $\mu = 0.04$.}
\end{figure}

If $f$ is assumed to have no noise, another way to perform upscaling with a
trained BLADE regularizer is to project the PDE as
\begin{equation}\label{e:blade-projected-upscaling}
\Hat{\vv{u}}^{(k+1)} = \Hat{\vv{u}}^{(k)} +
\dt P_0\bigl(\operatorname{BLADE}(\Hat{\vv{u}}^{(k)})\bigr),
\end{equation}
where, as detailed for instance in \cite{getreuer2011roussos},
$\Hat{\vv{u}}^{(0)}$ is constructed in the Fourier domain to satisfy
$\vv{f} = {\downarrow}(\varphi * \Hat{\vv{u}}^{(0)})$
and $P_0$ denotes orthogonal projection onto the subspace
$\{\vv{u} : {\downarrow}(\varphi * \vv{u}) = 0\}$.

\subsection{Segmentation}

BLADE TV is also useful for image segmentation. In the Chan--Vese ``active contours without edges''
segmentation method~\cite{chan2001active}, the segmentation contour is found as the zero level set of a function $\varphi$ that minimizes
\begin{equation}\label{e:chan-vese}
\begin{aligned}
&\mu \int \delta_\epsilon\bigl(\varphi(x)\bigr) \lvert\nabla\varphi\rvert
+ \nu \int H_\epsilon\bigl(\varphi(x)\bigr) \\
&{+}\: \lambda_1 \int\lvert f(x) - c_1 \rvert^2 H_\epsilon\bigl(\varphi(x)\bigr) \\
&{+}\: \lambda_2 \int \lvert f(x) - c_2 \rvert^2 \bigl(1 - H_\epsilon\bigl(\varphi(x)\bigr)\bigr),
\end{aligned}
\end{equation}
where $c_1$, $c_2$ are scalars that are simultaneously optimized, $H_\epsilon$ denotes a smoothed version of the Heaviside step function, $\delta_\epsilon$ is its derivative, and $\mu$, $\nu$, $\lambda_1$, $\lambda_2$ are constant parameters. In the gradient descent equation for $\varphi$, a scaled total variation flow appears in the first term:
\begin{equation}\label{e:chan-vese-descent}
\begin{aligned}
\partial_t \varphi &=
\delta_\epsilon(\varphi)
\bigl[ \mu \operatorname{div}\bigl(\tfrac{\nabla\varphi}{\lvert\nabla\varphi\rvert}\bigr) \\
&\hphantom{=}{-}\: \nu - \lambda_1 (f - c_1)^2 + \lambda_2 (f - c_2)^2 \bigr].
\end{aligned}
\end{equation}
We insert BLADE TV flow from the previous section to approximate Chan--Vese gradient descent as
\begin{equation}\label{e:blade-cv}
\begin{aligned}
\varphi^{(k+1)} &= \varphi^{(k)} +
\dt \, \delta_\epsilon(\varphi^{(k)})
\bigl[ \mu \operatorname{BLADE}(\varphi^{(k)}) \\
&\hphantom{=}{-}\: \nu - \lambda_1 (f - c_1^{(k)})^2 + \lambda_2 (f - c_2^{(k)})^2 \bigr].
\end{aligned}
\end{equation}

Figure~\ref{f:cv-wrench} shows an example segmentation with (\ref{e:blade-cv}) with $\mu = 0.04$, $\nu = 0$, $\lambda_1 = \lambda_2 = 1$, $\mathit{dt} = 0.5$ at convergence after 300 iterations, and compares with a reference implementation~\cite{getreuer2012chan} of
Chan--Vese, starting from a checkerboard initialization as suggested in~\cite{chan2001active}. When evolved for such a long time, the BLADE approximation is well beyond the 10 frames of unrolling done during training, and the approximation is blurrier than the reference. To counteract this effect, the BLADE result shown was performed with $\mu = 0.04$ while the reference was with $\mu = 0.2$ and otherwise same parameters. With this adjustment, the BLADE-based segmentation is qualitatively similar,
capturing nearly the same boundary for around the wrench, including the small
hole in the handle.

\begin{figure}[htb]
\centering
\begin{tabular}{@{}c@{\,}c@{\,}c@{}}
\small Reference implementation & \small BLADE implementation \\
\includegraphics[width=0.4\textwidth]{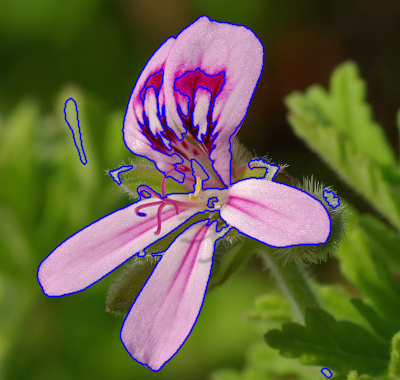} &
\includegraphics[width=0.4\textwidth]{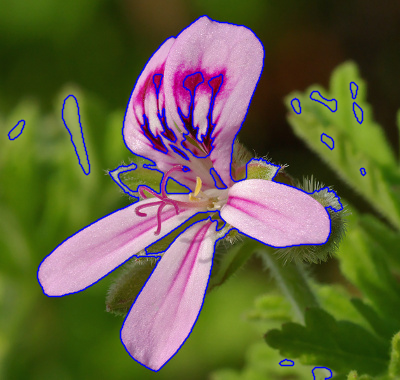}
\end{tabular}
\caption{\label{f:cvs-flower}Example Chan--Vese--Sandberg segmentation. Left: segmentation with reference implementation, $\mu = 0.2$. Right:
segmentation with BLADE TV approximation, $\mu = 0.04$.}
\end{figure}

For color image segmentation, the Chan--Vese--Sandberg method extends (\ref{e:chan-vese-descent}) to
\begin{equation}\label{e:chan-vese-sandberg}
\begin{aligned}
\partial_t \varphi &=
\delta_\epsilon(\varphi)
\bigl[ \mu \operatorname{div}\bigl(\tfrac{\nabla\varphi}{\lvert\nabla\varphi\rvert}\bigr) \\
&\hphantom{=}{-}\: \nu - \lambda_1 \lVert f - c_1 \rVert^2 + \lambda_2 \lVert f - c_2 \rVert^2 \bigr],
\end{aligned}
\end{equation}
in which $f$ is now a color image and $c_1$, $c_2$ are color vectors, yet $\varphi$ and the TV flow term have a single channel as before. We approximate Chan--Vese--Sandberg segmentation by again replacing the TV flow term with BLADE TV.
Figure~\ref{f:cvs-flower} shows an example segmentation.

\section{Optical Flow}

\begin{figure}[h]
\centering
\mbox{\beginpgfgraphicnamed{images/blade-linearized-bce-model}%

\begin{tikzpicture}[>=latex]

\node (Bladex) at (0,0.32) [draw=black,fill=PlotColorB!50,rectangle,rounded corners=2pt,inner sep=4pt] {\small$\operatorname{BLADE}^x$};
\node (Bladey) at (0,-0.32) [draw=black,fill=PlotColorB!50,rectangle,rounded corners=2pt,inner sep=4pt] {\small$\operatorname{BLADE}^y$};
\coordinate (below) at (0,-0.9);

\node (scalex) at (1.4,0.32) [draw=black,fill=gray!30,circle,inner sep=1pt] {\footnotesize $\delta^x$};

\node (scaley) at (1.4,-0.32) [draw=black,fill=gray!30,circle,inner sep=1pt] {\footnotesize $\delta^y$};

\node (adder) at (2.3,0) [draw=black,fill=gray!30,circle,inner sep=1pt] {\small\bf +};

\draw [rounded corners=2pt] 
(-4em,0) node [left] {$\smash{\vv{u}}\vphantom{x}$}
-- (-3em,0) -- (-3em,0.32) [->] -- (Bladex);
\draw [rounded corners=2pt] 
(-3em,0) -- (-3em,-0.32) [->] -- (Bladey);

\draw [->] (Bladex) -- (scalex);
\draw [->] (Bladey) -- (scaley);

\draw [rounded corners=2pt] 
(scalex) -- (1.85,0.32) [->] -- (adder);
\draw [rounded corners=2pt] 
(scaley) -- (1.85,-0.32) [->] -- (adder);

\draw [->] (adder) -- +(1.7em,0) node [right] {$\smash{\Tilde{\vv{u}}}\vphantom{x}$};

\draw [rounded corners=2pt] (-3.5em,0) -- +(0,-0.9) -- (below);
\draw [->,rounded corners=2pt] (below) -- (below-|adder) -- (adder);

\end{tikzpicture}%
\endpgfgraphicnamed}
\caption{\label{fig:blade-linearized-bce-model}
Network for resampling by a fractional displacement, based on BLADE approximation
of a version of the brightness constancy equation (\ref{e:linearized-bce}).}
\end{figure}
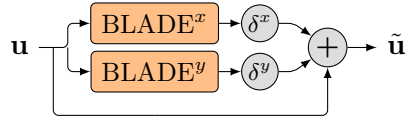

\begin{figure}[h]
\centering
\mbox{\beginpgfgraphicnamed{images/resampling-training-preparation}%
\input{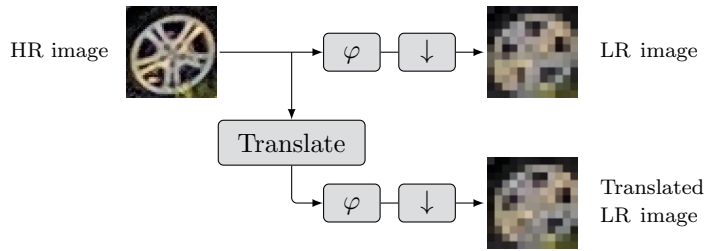}%
\endpgfgraphicnamed}
\caption{\label{fig:resampling-training-preparation}Preparing training data for BLADE-based resampling.}
\end{figure}

\begin{figure}[h]
\centering
\begin{tabular}{@{}c@{\,}c@{}}
\small Input (frame 0) & \small Exact (frame 2) \\
\includegraphics[width=0.4\textwidth]{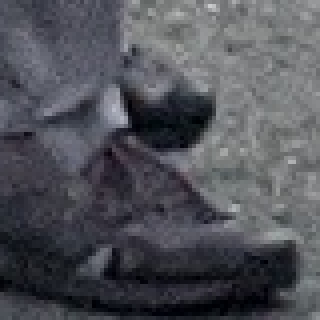} &
\includegraphics[width=0.4\textwidth]{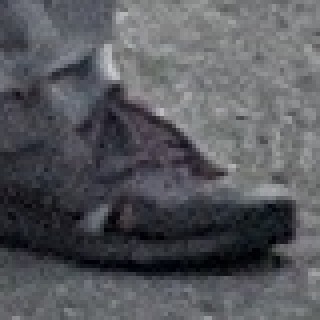} \\
\small Bicubic resampling & \small BLADE resampling \\
\includegraphics[width=0.4\textwidth]{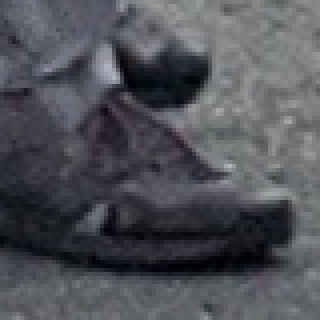} &
\includegraphics[width=0.4\textwidth]{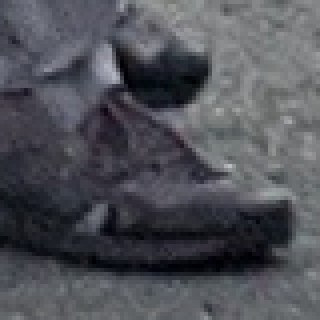}
\end{tabular}
\caption{\label{f:optical-flow-shoe}Comparison of bicubic and BLADE for resampling frame~0 to frame~1, then resampling that result from frame~1 to frame~2.}
\end{figure}

\begin{figure}[h]
\centering
\begin{tabular}{@{}c@{\,}c@{}}
\small Input (frame 0) & \small Exact (frame 2) \\
\includegraphics[width=0.4\textwidth]{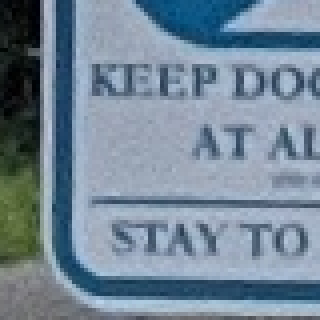} &
\includegraphics[width=0.4\textwidth]{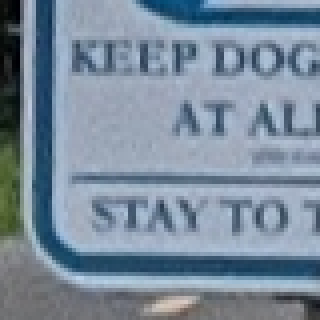} \\
\small Bicubic resampling & \small BLADE resampling \\
\includegraphics[width=0.4\textwidth]{images/sign-crop4x-frame_0_2_bicubic.png} &
\includegraphics[width=0.4\textwidth]{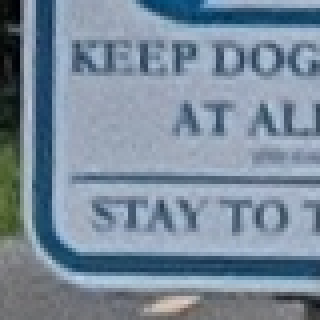}
\end{tabular}
\caption{\label{f:optical-flow-sign}Another example of resampling frame~0 to frame~1, then resampling frame~1 to frame~2.}
\end{figure}

In this section, we describe a method of resampling an image at displaced positions using BLADE. Given an image $\vv{u}$ and displacement field $(\vv{v}^x, \vv{v}^y)$, we seek to compute a resampled image whose $(m,n)$th pixel is sampled at the displaced location $(m + v^x_{m,n}, n + v^y_{m,n})$.
Such resampling is needed for instance in optical flow or image registration when mapping a moving image to a fixed image. This BLADE-based resampling is a variation of the RAISR super-resolution method of Romano et al.~\cite{romano2017raisr}. Super-resolution can be seen as a case of resampling at regularly-spaced sample locations; the new difficulty here is that the sampling locations are typically irregular as determined by the displacement field $(\vv{v}^x, \vv{v}^y)$.

We model the observed image $\vv{u}$ as having been sampled from a convolution of an underlying continuous domain image $U(x, y)$ with a point spread function $\varphi(x, y)$, plus noise,
\begin{equation}
u_{m,n} = (\varphi * U)(m, n) + \mathit{noise}_{m, n}.
\end{equation}
The desired resampled image, displaced by $(\vv{v}^x, \vv{v}^y)$, is
\begin{equation}\label{e:resampled-model}
\Tilde{u}_{m,n} = (\varphi * U)(m + v^x_{m,n}, n + v^y_{m,n}).
\end{equation}

Considering a fixed pixel $(m,n)$, we decompose the displacement vector $(v^x_{m,n}, v^y_{m,n})$ into integer and fractional parts,
\begin{equation}
v^x = [ v^x ] + \delta^x, \quad v^y = [ v^y ] + \delta^y
\end{equation}
where $[ \cdot ]$ denotes rounding to the nearest integer and $\delta^x, \delta^y$ are fractional pixel displacements in $[-1/2, +1/2]$. For the integer part, (\ref{e:resampled-model}) reduces to indexing a denoised estimate of $\vv{u}$ at a position shifted by a whole number of pixels. Therefore, we focus on the fractional part.

For a fractional displacement, we have by Taylor expansion
\begin{align}
\Tilde{u}_{m,n} &= (\varphi * U)(m + \delta^x, n + \delta^y) \nonumber \\
&\approx (\varphi * U)(m, n) + \delta^x (\partial_x \varphi * U)
+ \delta^y (\partial_x \varphi * U) \nonumber \\
&= \bigl((\varphi + \delta^x \partial_x \varphi + \delta^y \partial_y \varphi) * U\bigr)(m, n) \nonumber \\
&= \bigl((I + \delta \cdot \grad) \varphi * U)\bigr)(m, n).
\label{e:linearized-bce}
\end{align}
Therefore in this sense, resampling (\ref{e:resampled-model}) corresponds to applying the differential filter $(I + \delta \cdot \grad)$. This is essentially linearization of a version of the brightness constancy equation or the optical flow constraint equation including a point spread function.

We approximate (\ref{e:linearized-bce}) with BLADE as
\begin{align}
\Tilde{\vv{u}} = \vv{u} + \delta^x \operatorname{BLADE}^x(\vv{u})
 + \delta^y \operatorname{BLADE}^y(\vv{u}),
\end{align}
in which $\operatorname{BLADE}^x$ and $\operatorname{BLADE}^y$ have independent filters (Figure~\ref{fig:blade-linearized-bce-model}).
We prepare data for training by beginning with a high-resolution image, translating, convolving with $\varphi$, and downsampling (Figure~\ref{fig:resampling-training-preparation}). By this process we create example pairs of an observed (low resolution) image and a target image translated by a subpixel displacement. For the point spread function $\varphi$, we use a Gaussian with standard deviation of $0.5$ pixels.

We test the method on image bursts from a handheld camera so that there is significant motion (Figures~\ref{f:optical-flow-shoe} and \ref{f:optical-flow-sign}). The flow field between frames was estimated by a pyramid-based block matching alignment algorithm as described in~\cite{wronski2019handheld}. To emphasize the effect of resampling, we show the outcome after two steps of resampling: frame~0 is resampled to frame~1, then that result is resampled again to frame~2. Resampling with standard bicubic is also shown. In comparison, the BLADE result is sharper and captures the structure better.

\section{Conclusions and Future Work}

We have shown that BLADE, a shallow 2-layer network, is reliable and efficient
in approximating several hyperbolic PDEs in image processing. The use of machine learning for approximating PDEs raises a number of questions. For instance, stability is an essential property in PDE methods. An interesting question is whether it is possible to develop learning-based PDE methods that are both provably stable and have flexible capacity to learn.

\bibliographystyle{siamplain}
\bibliography{article}

\begin{thebibliography}{10}

\bibitem{abadi2016tensorflow}
{\sc M.~Abadi, P.~Barham, J.~Chen, Z.~Chen, A.~Davis, J.~Dean, M.~Devin,
  S.~Ghemawat, G.~Irving, M.~Isard, et~al.}, {\em {Tensorflow:} a system for
  large-scale machine learning}, in 12th USENIX Symposium on Operating Systems
  Design and Implementation (OSDI 16), 2016, pp.~265--283.

\bibitem{andreu2001minimizing}
{\sc F.~Andreu, C.~Ballester, V.~Caselles, J.~M. Maz{\'o}n, et~al.}, {\em
  Minimizing total variation flow}, Differential and integral equations, 14
  (2001), pp.~321--360.

\bibitem{bertozzi2007inpainting}
{\sc A.~L. Bertozzi, S.~Esedoḡlu, and A.~Gillette}, {\em Inpainting of binary
  images using the {Cahn--Hilliard} equation}, IEEE Transactions on Image
  Processing, 16 (2007), pp.~285--291.

\bibitem{chan2001active}
{\sc T.~F. Chan and L.~A. Vese}, {\em Active contours without edges}, IEEE
  Transactions on Image Processing, 10 (2001), pp.~266--277.

\bibitem{chang2017multi}
{\sc B.~Chang, L.~Meng, E.~Haber, F.~Tung, and D.~Begert}, {\em Multi-level
  residual networks from dynamical systems view}, arXiv preprint
  arXiv:1710.10348,  (2017).

\bibitem{chen2018neural}
{\sc R.~T. Chen, Y.~Rubanova, J.~Bettencourt, and D.~Duvenaud}, {\em Neural
  ordinary differential equations}, arXiv preprint arXiv:1806.07366,  (2018).

\bibitem{choi2018fast}
{\sc S.~Choi, J.~Isidoro, P.~Getreuer, and P.~Milanfar}, {\em Fast, trainable,
  multiscale denoising}, in 2018 25th IEEE International Conference on Image
  Processing (ICIP), IEEE, 2018, pp.~963--967.

\bibitem{esedoglu2001analysis}
{\sc S.~Esedoḡlu}, {\em An analysis of the {Perona--Malik} scheme},
  Communications on Pure and Applied Mathematics: A Journal Issued by the
  Courant Institute of Mathematical Sciences, 54 (2001), pp.~1442--1487.

\bibitem{evans2010}
{\sc L.~C. Evans}, {\em Partial differential equations}, American Mathematical
  Society, Providence, R.I., 2010.

\bibitem{getreuer2011roussos}
{\sc P.~Getreuer}, {\em {Roussos--Maragos} tensor-driven diffusion for image
  interpolation}, {Image Processing On Line}, 1 (2011), pp.~178--186,
  \url{https://doi.org/10.5201/ipol.2011.g_rmdi}.

\bibitem{getreuer2012chan}
{\sc P.~Getreuer}, {\em {Chan--Vese} segmentation}, {Image Processing On Line},
  2 (2012), pp.~214--224.
\newblock \url{https://doi.org/10.5201/ipol.2012.g-cv}.

\bibitem{getreuer2018blade}
{\sc P.~Getreuer, I.~Garcia-Dorado, J.~Isidoro, S.~Choi, F.~Ong, and
  P.~Milanfar}, {\em {BLADE:} filter learning for general purpose computational
  photography}, in 2018 IEEE International Conference on Computational
  Photography (ICCP), IEEE, 2018, pp.~1--11,
  \url{https://doi.org/10.1109/ICCPHOT.2018.8368476}.

\bibitem{han2018solving}
{\sc J.~Han, A.~Jentzen, and E.~Weinan}, {\em Solving high-dimensional partial
  differential equations using deep learning}, Proceedings of the National
  Academy of Sciences, 115 (2018), pp.~8505--8510.

\bibitem{he2016deep}
{\sc K.~He, X.~Zhang, S.~Ren, and J.~Sun}, {\em Deep residual learning for
  image recognition}, in Proceedings of the IEEE conference on computer vision
  and pattern recognition, 2016, pp.~770--778.

\bibitem{johnson2016perceptual}
{\sc J.~Johnson, A.~Alahi, and L.~Fei-Fei}, {\em Perceptual losses for
  real-time style transfer and super-resolution}, in European Conference on
  Computer Vision, Springer, 2016, pp.~694--711.

\bibitem{lele1992compact}
{\sc S.~Lele}, {\em Compact finite difference schemes with spectral-like
  resolution.}, Journal of Computational Physics, 103 (1992), pp.~16--42,
  \url{https://doi.org/10.1016/0021-9991(92)90324-R}.

\bibitem{long2018pde}
{\sc Z.~Long, Y.~Lu, X.~Ma, and B.~Dong}, {\em Pde-net: Learning pdes from
  data}, in International Conference on Machine Learning, PMLR, 2018,
  pp.~3208--3216.

\bibitem{lu2018beyond}
{\sc Y.~Lu, A.~Zhong, Q.~Li, and B.~Dong}, {\em Beyond finite layer neural
  networks: Bridging deep architectures and numerical differential equations},
  in International Conference on Machine Learning, PMLR, 2018, pp.~3276--3285.

\bibitem{perona1990scale}
{\sc P.~Perona and J.~Malik}, {\em Scale-space and edge detection using
  anisotropic diffusion}, IEEE Transactions on Pattern Analysis and Machine
  Intelligence, 12 (1990), pp.~629--639.

\bibitem{rodrigues2018deepdownscale}
{\sc E.~R. Rodrigues, I.~Oliveira, R.~L. Cunha, and M.~A. Netto}, {\em
  {DeepDownscale}: a deep learning strategy for high-resolution weather
  forecast}, arXiv preprint arXiv:1808.05264,  (2018).

\bibitem{romano2017raisr}
{\sc Y.~Romano, J.~Isidoro, and P.~Milanfar}, {\em {RAISR:} rapid and accurate
  image super resolution}, IEEE Transactions on Computational Imaging, 3
  (2017), pp.~110--125.

\bibitem{rudin1992nonlinear}
{\sc L.~I. Rudin, S.~Osher, and E.~Fatemi}, {\em Nonlinear total variation
  based noise removal algorithms}, Physica D: Nonlinear Phenomena, 60 (1992),
  pp.~259--268.

\bibitem{ruthotto2019deep}
{\sc L.~Ruthotto and E.~Haber}, {\em Deep neural networks motivated by partial
  differential equations}, Journal of Mathematical Imaging and Vision,  (2019),
  pp.~1--13.

\bibitem{sorteberg2018approximating}
{\sc W.~E. Sorteberg, S.~Garasto, A.~S. Pouplin, C.~D. Cantwell, and A.~A.
  Bharath}, {\em Approximating the solution to wave propagation using deep
  neural networks}, arXiv preprint arXiv:1812.01609,  (2018).

\bibitem{talebi2018learned}
{\sc H.~Talebi and P.~Milanfar}, {\em Learned perceptual image enhancement}, in
  Computational Photography (ICCP), 2018 IEEE International Conference on,
  IEEE, 2018, pp.~1--13.

\bibitem{wang2004image}
{\sc Z.~Wang, A.~C. Bovik, H.~R. Sheikh, and E.~P. Simoncelli}, {\em Image
  quality assessment: from error visibility to structural similarity}, IEEE
  Transactions on Image Processing, 13 (2004), pp.~600--612.

\bibitem{weickert1998anisotropic}
{\sc J.~Weickert}, {\em Anisotropic diffusion in image processing}, vol.~1,
  Teubner Stuttgart, 1998.

\bibitem{weickert2002scheme}
{\sc J.~Weickert and H.~Scharr}, {\em A scheme for coherence-enhancing
  diffusion filtering with optimized rotation invariance}, Journal of Visual
  Communication and Image Representation, 13 (2002), pp.~103--118.

\bibitem{weinan2017proposal}
{\sc E.~Weinan}, {\em A proposal on machine learning via dynamical systems},
  Communications in Mathematics and Statistics, 5 (2017), pp.~1--11.

\bibitem{weinan2018deep}
{\sc E.~Weinan and B.~Yu}, {\em The deep {Ritz} method: a deep learning-based
  numerical algorithm for solving variational problems}, Communications in
  Mathematics and Statistics, 6 (2018), pp.~1--12.

\bibitem{wronski2019handheld}
{\sc B.~Wronski, I.~Garcia-Dorado, M.~Ernst, D.~Kelly, M.~Krainin, C.-K. Liang,
  M.~Levoy, and P.~Milanfar}, {\em Handheld multi-frame super-resolution}, ACM
  Transactions on Graphics (TOG), 38 (2019), pp.~1--18.

\end{thebibliography}

\end{document}